\documentclass[runningheads]{llncs}
\usepackage[T1]{fontenc}
\usepackage{graphicx,verbatim, amsmath, amssymb}
\usepackage{algorithm}
\usepackage{algorithmic}
\usepackage{booktabs}
\usepackage{hyperref}
\usepackage{siunitx}

\usepackage{color}
\begin{document}
\title{Low-Rank-Modulated Functa: Exploring the Latent Space of Implicit Neural Representations for Interpretable Ultrasound Video Analysis }
\titlerunning{Low-Rank-Modulated Functa}
%
\author{Julia Wolleb\inst{1,2}\and
Cristiana Baloescu\inst{3} \and
Alicia Durrer \inst{4}
Hemant D. Tagare \inst{5} \and
Xenophon Papademetris \inst{1,2,5}}
\authorrunning{J. Wolleb et al.}
%

\authorrunning{J. Wolleb et al.}

\institute{Dept. of Biomedical Informatics \& Data Science, Yale University, New Haven, USA \and
Yale Biomedical Imaging Institute, Yale University, New Haven, USA \and
Department of Emergency Medicine, Yale School of
Medicine, New Haven,  USA \and
Dept. of Biomedical Engineering, University of Basel, Allschwil, Switzerland \and
Dept. of Radiology \& Biomedical Imaging, Yale University, New Haven, USA \\
\email{julia.wolleb@yale.edu}
}

\maketitle              
\begin{abstract}

Implicit neural representations (INRs) have emerged as a powerful framework for continuous image representation learning. In \textit{Functa}-based approaches, each image is encoded as a latent modulation vector that conditions a shared INR, enabling strong reconstruction performance. However, the structure and interpretability of the corresponding latent spaces remain largely unexplored.
In this work, we investigate the latent space of \textit{Functa}-based models for ultrasound videos and propose \textit{Low-Rank-Modulated Functa} (\textit{LRM-Functa}), a novel architecture that enforces a low-rank adaptation of modulation vectors in the time-resolved latent space. When applied to cardiac ultrasound, the resulting latent space exhibits clearly structured periodic trajectories, facilitating visualization and interpretability of temporal patterns. The latent space can be traversed to sample novel frames, revealing smooth transitions along the cardiac cycle, and enabling direct readout of end-diastolic (ED) and end-systolic (ES) frames without additional model training.
We show that \textit{LRM-Functa} outperforms prior methods in unsupervised ED and ES frame detection, while compressing each video frame to as low as rank $k=2$ without sacrificing competitive downstream performance on ejection fraction prediction. Evaluations on out-of-distribution frame selection in a cardiac point-of-care dataset, as well as on lung ultrasound for B-line classification, demonstrate the generalizability of our approach.
Overall, \textit{LRM-Functa} provides a compact, interpretable, and generalizable framework for ultrasound video analysis. The code is available at \url{https://github.com/JuliaWolleb/LRM_Functa}.


\end{abstract}
\section{Introduction}

\begin{figure}[t]
\includegraphics[width=\textwidth]{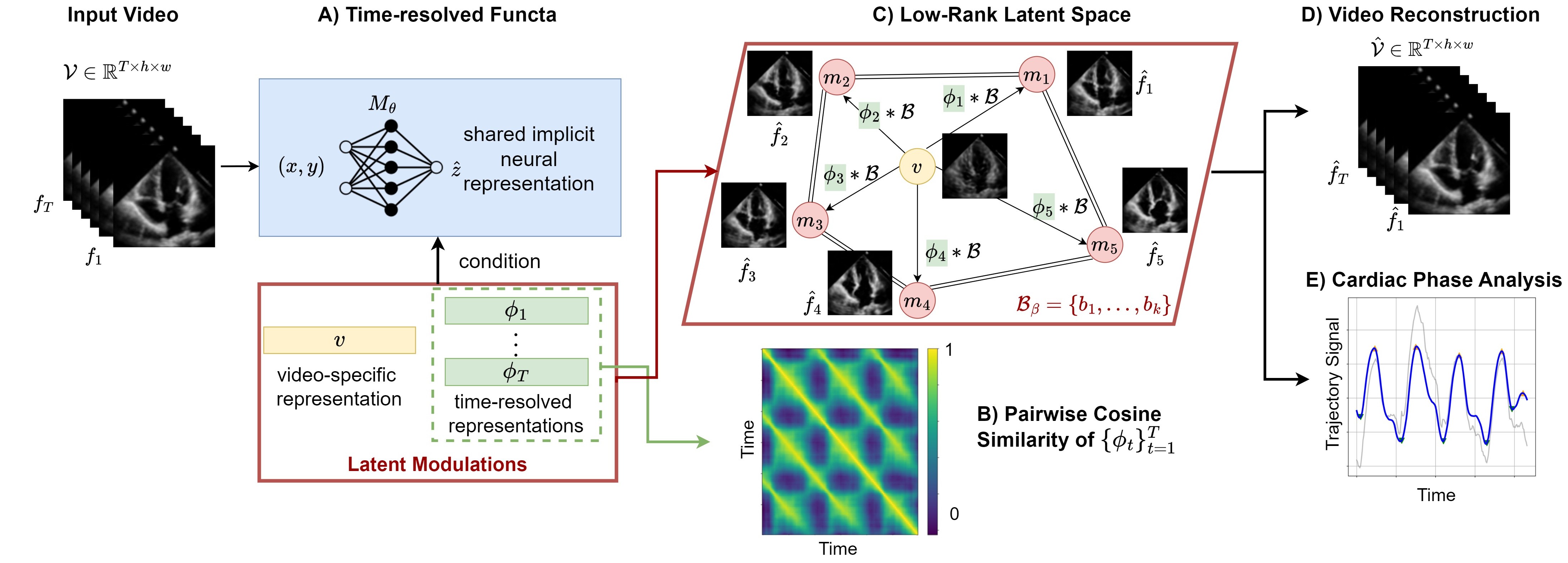}
\caption{Overview of our \textit{LRM-Functa} framework. A) Our approach builds on \textit{VidFuncta}, where each ultrasound video is represented by a video-level modulation vector $v$ and time-resolved modulation vectors $\phi=\{\phi_t\}_{t=1}^T$. (B) Pairwise cosine similarity of $\phi$ over time reveals a clear low-rank pattern. (C) Motivated by this observation, we constrain the latent modulation space to a low-rank subspace $\mathcal{B}_{\beta}$. (D/E) This structured latent space enables efficient compression-reconstruction and  cardiac phase analysis.} \label{basis}
\end{figure}

Ultrasound imaging is widely used in clinical practice due to its portability, low cost, and real-time acquisition. Modern workflows rely on ultrasound videos rather than single frames, particularly in cardiac and lung imaging, where temporal dynamics are critical for diagnosis and quantification \cite{al2024spatiotemporal,hernandez2021deep,takizawa2025videoclipmodelmultiview}. 
Precise identification of end-systolic (ES) and end-diastolic (ED) frames is essential in echocardiography because key measures, such as ejection fraction and ventricular hypertrophy, are defined at specific cardiac phases \cite{mada2015define,oikonomou2024artificial}.
However, ultrasound videos present challenges for storage, transmission, and analysis due to their size, noise, and high temporal redundancy. This has motivated the development of low-dimensional and task-specific video representations \cite{jiao2020self,jin2025artificial}. 
In this work, we build on implicit neural representations (INRs) \cite{sitzmann2020implicit} to learn compact, continuous representations for ultrasound video compression and analysis. We adopt the \textit{Functa} framework \cite{dupont2022data,friedrichmedfuncta}, which encodes entire datasets using a shared INR backbone modulated by per-image latent vectors. While this approach achieves efficient reconstruction, the structure and interpretability of the latent space remains largely unexplored.
To address this gap, building on time-resolved \textit{VidFuncta} \cite{wolleb2025vidfuncta} and inspired by \cite{yang2025latent}, we propose \textit{Low-Rank-Modulated Functa (LRM-Functa)}, shown in Figure \ref{basis}. 
This novel architecture enforces a structured, low-rank latent space shown in Figure \ref{basis}.C, producing compact video representations at high compression rates. The latent trajectories are directly interpretable and can be visualized for unsupervised cardiac phase analysis, enabling extraction of ED and ES frame indices without additional model training.

\subsubsection{Related Work}

INRs model signals as continuous neural functions and have been applied to 2D, 3D, and spatiotemporal data \cite{zhu2025implicit}, where a neural network learns coordinate-to-intensity mappings for self-supervised reconstruction.
They are widely used in medical image analysis \cite{molaei2023implicit}, including ultrasound 3D reconstruction \cite{grutman2025implicit} or speed-of-sound estimation \cite{byra2024implicit}. The \textit{Functa} framework \cite{dupont2022data,friedrichmedfuncta} extends the INR paradigm by combining a shared network $M_{\theta}$ with per-image latent modulation vectors, enabling entire datasets to be encoded within a single model and allowing efficient  reconstruction \cite{bauer2023spatial,dupont2022data,dupont2022coin++,friedrichmedfuncta}. \textit{VidFuncta} \cite{wolleb2025vidfuncta} further extends this approach to videos through time-resolved modulation vectors, while Latent-INRs use frame-specific latent codes learned jointly with hypernetworks for video compression \cite{maiya2024latent}. However, these methods focus on reconstruction and downstream tasks, leaving the interpretability of the latent space largely unexplored. Structured latent representations have been explored in dynamic human view synthesis \cite{peng2021neural}. Low-rank adaptation methods (LoRA), originally developed for parameter-efficient fine-tuning of large language models \cite{hu2022lora}, have been applied to INRs for compact video compression \cite{truong2025low}. The paper \cite{mcginnis2025optimizing} analyzed how rank collapse in INR layers affects reconstruction quality. In medical image analysis, learned video representations support tasks such as cardiac phase detection, functional quantification, and pathology classification \cite{farhad2023cardiac,jiao2020self}. Prior work has investigated interpretable latent trajectories using variational and convolutional autoencoders to analyze cardiac motion \cite{ryser2022anomaly,yang2025latent}. However, structured and interpretable latent representations remain unexplored in INR-based architectures.

\subsubsection{Contributions}
To the best of our knowledge, we are the first to analyze and structure the latent space of \textit{Functa}-based frameworks. Our novel architecture, \textit{LRM-Functa}, enforces a low-rank latent representation, enabling interpretable visualization and analysis of temporal trajectories in cardiac ultrasound. This enables direct identification of ED and ES frames from the latent trajectories without additional model training, outperforming state-of-the-art unsupervised methods, and generalizing to out-of-distribution (OOD) cardiac views. Furthermore, we explore the compression–reconstruction trade-off, demonstrating that accurate ejection fraction estimation is preserved even under high compression down to $k=2$ components per frame. We additionally validate the generalizability of our approach on B-line detection in lung ultrasound (LUS) videos.

\section{Method}

\subsection{Exploration of the Latent Space of VidFuncta}

We explore the \textit{VidFuncta} architecture as our baseline INR for ultrasound video analysis \cite{wolleb2025vidfuncta}, illustrated in Figure \ref{basis}.A. \textit{VidFuncta} employs a shared neural network $M_{\theta}$, implemented as a multilayer perceptron (MLP), whose learnable parameters $\theta$ are shared across the whole dataset. Each video is conditioned on a learnable video-level latent vector 
$v$ using FiLM-style modulation \cite{perez2018film}, and temporal dynamics are captured by frame-specific modulation vectors 
$\phi_t$, one per frame $f_t$ for 
$t=1,…,T$. To explore the structure of this latent space, we visualize pairwise cosine similarities between all frame vectors $\phi = \{\phi_t\}_{t=1}^T$. The resulting 
$T \times T$ matrices exhibit low-rank patterns with clear periodicity along the side diagonals, as shown in Figure \ref{basis}.B, indicating that the latent space encodes rich temporal dynamics. 
This observation motivates the low-rank constraint on the latent space we propose in this work, as described in Section \ref{sec:lora}.

\subsection{Low-Rank Adaptation of the Latent Modulation Vectors} \label{sec:lora}

\begin{figure}[h]
\center
\includegraphics[width=1\textwidth]{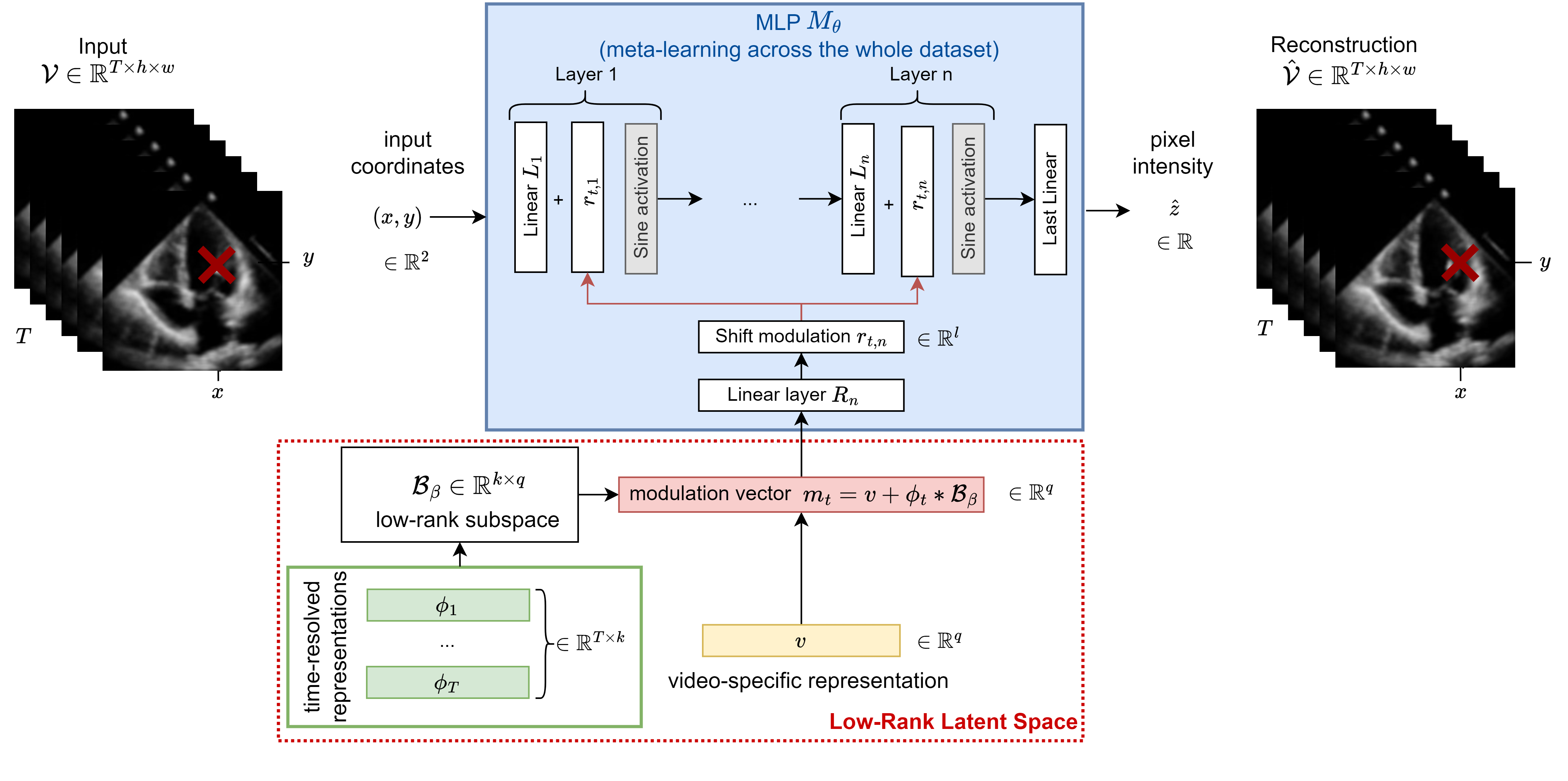}
\caption{Proposed architecture for low-rank adaptation of the modulation vectors. Each video $\mathcal{V}$ is encoded into a video-level representation $v$ and time-resolved frame representations $\phi_t$, which are projected onto a rank-$k$ subspace $\mathcal{B}_{\beta}$ to produce modulation vectors $m_t = v + \mathcal{B}_{\beta}  \phi_t$. These vectors shift the activations of a shared MLP $M_\theta$, trained to reconstruct pixel intensities 
 $\hat{z}$ 
 at each coordinate 
$(x,y)$, shown as a red cross.} \label{architecture}
\end{figure}

We present \textit{LRM-Functa}, which defines a low-rank latent space within the \textit{VidFuncta} framework for video representation. Building on LoRA \cite{hu2022lora,truong2025low} and latent motion profiling in convolutional autoencoders \cite{yang2025latent}, each frame-specific modulation vector is constrained to $m_t = v + \mathcal{B}_{\beta} \, \phi_t $, as illustrated in Figure \ref{basis}.C. Here, $v \in \mathbb{R}^q$ denotes the video-level latent vector, $\mathcal{B}_{\beta} = \{b_1, \dots, b_k\} \in \mathbb{R}^{k \times q}$ defines a learnable rank-$k$ subspace of $\mathbb{R}^q$, and $\phi_t \in \mathbb{R}^k$ represents the learnable frame-specific update coefficients. This ensures that all updates from $v$ to individual frames $f_t$ are constrained to the low-rank subspace spanned by $\mathcal{B}_{\beta}$. If rank $k=2$, $\mathcal{B}_{\beta}$ can be visualized as a 2D plane. The architecture is shown in Figure \ref{architecture}.

\paragraph{Orthogonal Constraint:} To explore the structure of the low-rank subspace $\mathcal{B}_{\beta} \in \mathbb{R}^{k\times q}$ with learnable parameters $\beta$, we study two variants: The baseline \textit{LRM-Functa basic} (\textit{LRM-Functa}$_{b}$) initializes $\beta$ randomly, whereas \textit{LRM-Functa ortho} (\textit{LRM-Functa}$_{o}$) uses an orthogonal basis initialization. During training, the modulation vectors $v$ and $\phi$, and the shared parameters $\theta$ and $\beta$, are updated following the meta-learning schedule used in \cite{wolleb2025vidfuncta}.
The loss function is given by 
\begin{equation}
    \mathcal{L}_{t} =\frac{1}{N} \sum_{i=1}^N \lVert M_{\theta, v, \phi_t, \beta}(x_{i},y_{i} ) - z_i \rVert _2^2  + \lambda_{ortho} \big\|\mathcal{B}_{\beta}^\top \mathcal{B}_{\beta} - I \big\|_F,
\end{equation}
where $\| \cdot\|_F$ denotes the Frobenius norm, $I$ is the identity matrix, and $\lambda_{ortho}=0$ in the basic case, and  $\lambda_{ortho}=1$ in the orthogonal variant.

\subsection{Inference and Downstream Task Integration}
\label{inference}

          

\begin{figure} [t]
\includegraphics[width=1\textwidth]{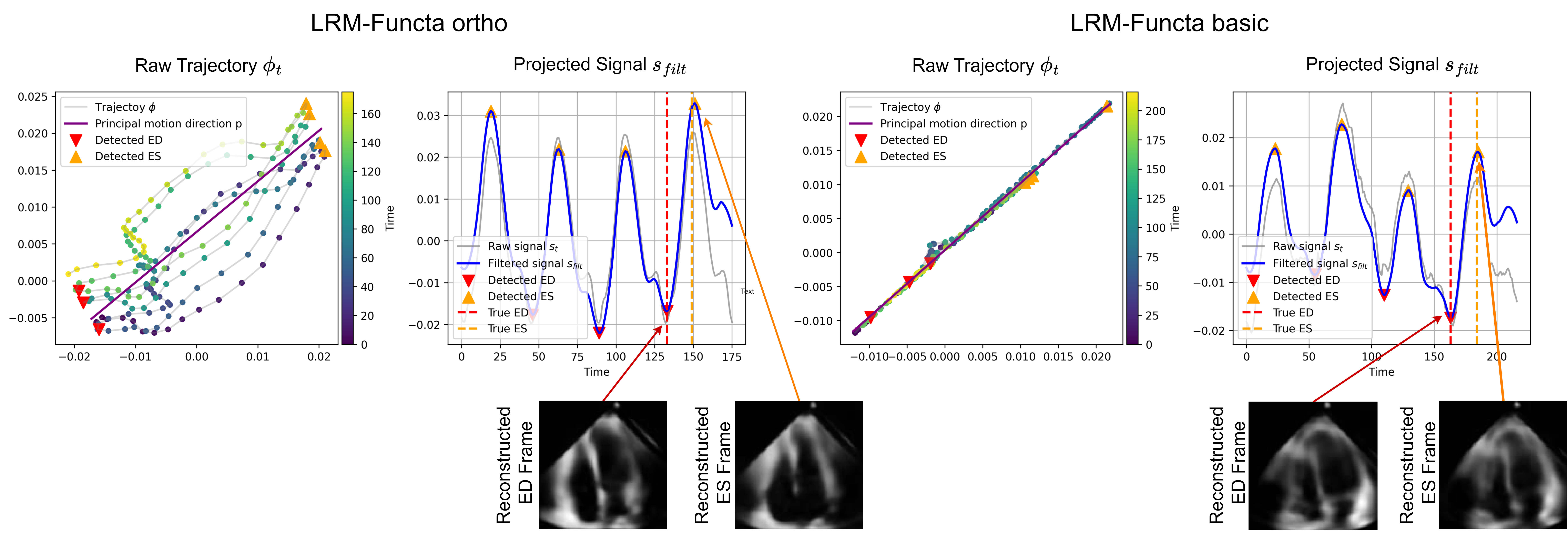}
\caption{
For $k = 2$, we visualize the raw trajectories $\phi_t$ over time. \textit{LRM-Functa}$_o$ exhibits spiral-like trajectories, whereas \textit{LRM-Functa}$_b$ collapses the trajectory to a line. 
Projecting $\phi_t$ onto the principal motion direction $p$ and filtering yields the 1D signal 
$s_{filt}$, which enables direct interpretable identification of ED and ES frames.} \label{trajectories}
\mbox{ }
\mbox{ }
    \centering
    \includegraphics[width=1\linewidth]{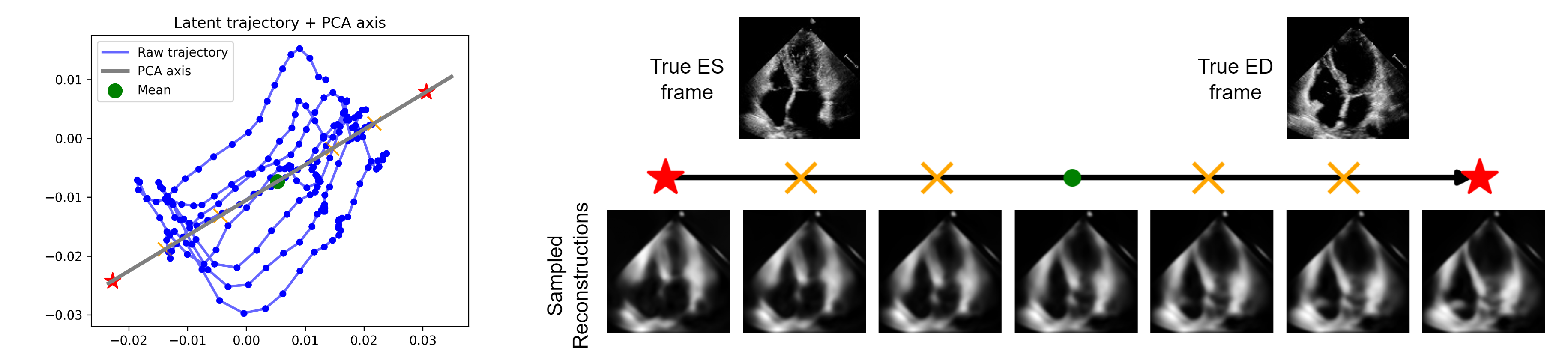}
    \caption{We plot the raw trajectory $\phi$, its mean, and and first principal component for using \textit{LRM-Functa}$_{o}$.
    Walking along this axis generates latent encodings, from which reconstructed frames show a smooth progression from ES to ED. Overshooting, indicated by red stars, produces exaggerated contraction and relaxation. }
    \label{fig:sampling}
\end{figure}

During inference, we follow Algorithm 1 of \textit{VidFuncta} \cite{wolleb2025vidfuncta}, freezing the shared parameters $\theta$ and $\beta$ while fitting the modulation vectors $v$ and temporal modulations $\phi = \{\phi_t\}_{t=1}^T$ to each video. The fitted $v$ and $\phi$ are stored as a compact representation of the video, and the reconstructed video $\hat{\mathcal{V}}$ is computed. 
To extract signals from the raw trajectories $\phi$, we apply Algorithm 1 of \cite{yang2025latent},  as shown in Figure \ref{trajectories}. We estimate the principal motion direction $p$ using PCA \cite{wold1987principal} on the motion directions $d_t =\frac{\mathbf{\phi_{t+1} - \phi_{t}}}{\|\mathbf{\phi_{t+1} - \phi_{t}}\|}$. The raw signal $s_t$ is obtained by projecting $\phi_t$ onto $p$, which is corrected for baseline wander and smoothed via a Savitzky–Golay filter \cite{schafer2011savitzky} to obtain $s_{filt}$. ED and ES indices are identified as valleys and peaks of $s_{filt}$ using a prominence-based threshold.

\section{Experiments}
We consider three datasets. EchoNet-Dynamic \cite{ouyang2019echonet} contains 10,030 four-chamber cardiac videos labeled with ejection fraction values and ES/ED frames. For OOD evaluation, we use 17 point-of-care cardiac ultrasound (POCUS) videos showing a two-chamber parasternal long axis view, collected and labeled for ES/ED frames at \textit{anonymous hospital}. We further considered 200 lung ultrasound video clips from adult emergency department patients at \textit{anonymous hospital}, each annotated for the presence of B-lines. We reserved 40 cases for testing. Ethics approval was obtained for the creation of these retrospective, anonymized datasets. All videos are downsampled to a spatial resolution of $112\times 112$, and normalized to values between 0 and 1. 
We use Python version 3.11.13 and PyTorch version 2.4.1. The vector $v$ has dimension $q=256$, and we vary $k$ between $2$ and $512$.  All remaining hyperparameters follow the configurations of \cite{wolleb2025vidfuncta}. All \textit{LRM-Functa} models are trained for 80,000 iterations on a 24GB NVIDIA RTX A5000 GPU, which takes 20 hours per model.  We compare \textit{LRM-Functa} against frame-wise \textit{COIN++} \cite{dupont2022coin++} and \textit{MedFuncta} \cite{friedrichmedfuncta}, time-resolved \textit{VidFuncta} \cite{wolleb2025vidfuncta}, and a convolutional \textit{Autoencoder} \cite{yang2025latent}.
We consider three tasks for quantitative anaylsis: \\   
 \textbf{Compression-Reconstruction Task:}
We evaluate reconstructions using Peak Signal-to-Noise Ratio (PSNR) and 3D Structural Similarity Index (SSIM3D) between the original video 
$ \mathcal{V}$ and its reconstruction $\hat{\mathcal{V}}$  on the predefined EchoNet-Dynamic test set of 1277 samples. Different compression rates are tested by varying the rank $k$ of the subspace $\mathcal{B}_{\beta}$, defining the length of the vectors $\phi_t$.
\\
\textbf{ED and ES Frame Detection:}
We apply the pipeline described in Section \ref{inference} to detect ED and ES frames on the trajectories $\phi$ on the EchoNet-Dynamic test set.  
We compute the mean absolute error (MAE) between the labeled frame and the closest detected frames.  
For OOD performance on the POCUS set, we use models pretrained on EchoNet-Dynamic train set without fine-tuning.
\\
 \textbf{Downstream Tasks on the Reconstructions:}
We first reconstruct the videos $\hat{\mathcal{V}}$ for EchoNet-Dynamic and the LUS dataset. On these reconstructions, a ResNet 18-3D \cite{hara2018can} is trained for 20 epochs using 5-fold cross-validation to predict ejection fraction and classify B-lines, respectively. Mean squared error (MSE) is used for ejection fraction prediction and binary cross-entropy loss for B-line classification. Performance is evaluated using MAE, $R^2$, and root mean squared error (RMSE) for ejection fraction prediction, and accuracy, F1-score, and area under the receiver operating characteristic curve (AUROC) for B-line classification.

\section{Results and Discussion}

\subsection{ Trajectory Visualization and Walk in the Latent Space}
As shown in Section \ref{quantitative}, we can compress $\phi_t$ to a length of $k = 2$ and visualize it directly. In Figure \ref{trajectories}, we show the raw latent trajectories $\phi = \{\phi_t\}_{t=1}^T$ over time for two EchoNethDynamic test examples.
\textit{LRM-Functa}$_{o}$ organizes the latent space into a periodic spiral, while in \textit{LRM-Functa}$_{b}$ the trajectories move back and forth  along a line, suggesting that even $k=1$ may be sufficient to capture cardiac motion. We project $\phi$ onto the principal motion direction $p$ and filter it to obtain $s_{filt}$, as described in Section \ref{inference}. Peaks and valleys of $s_{filt}$ allow unsupervised readout of ED and ES indices. 
Figure \ref{fig:sampling} shows the raw latent trajectory of $\phi$ for another test example using \textit{LRM-Functa}$_{o}$, with the first principal component of $\phi$ computed via PCA plotted as a gray axis. Traversing this latent axis and reconstructing frames from sampled points reveals a smooth ES-to-ED progression, demonstrating that the latent space encodes cardiac motion and supports continuous-time frame sampling.

\subsection{Quantitative Results}
\label{quantitative}
\subsubsection{Compression-Reconstruction Results:}
\begin{figure}[t]
\includegraphics[width=1\textwidth]{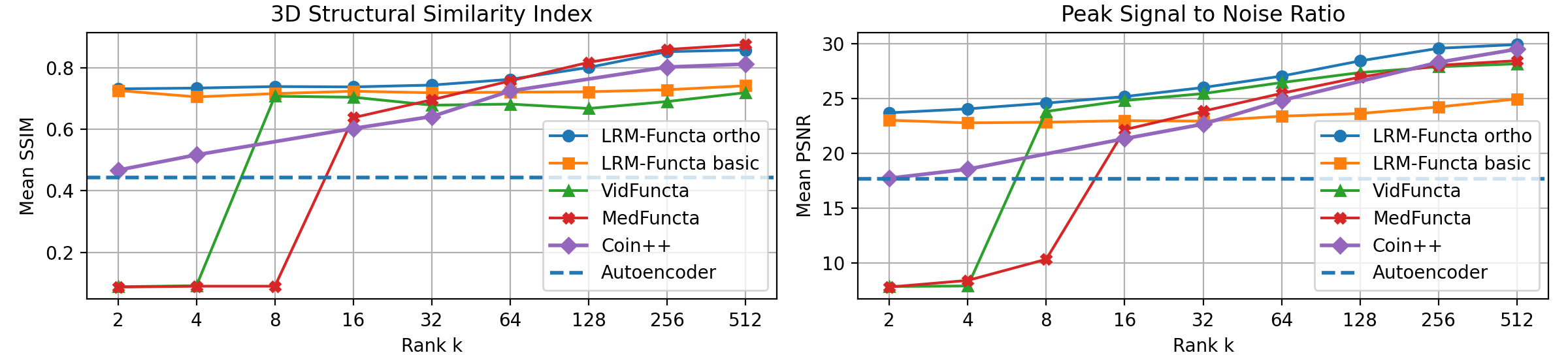}
\caption{Mean SSIM and PSNR scores on the EchoNet-Dynamic test set for all comparing methods across varying latent dimensions $k$ (x-axis shown on a log2
 scale). For small $k$, our \textit{LRM-Functa} variants are the only methods to produce stable reconstructions, while for higher ranks, \textit{LRM-Functa}$_{o}$  achieves among the highest performance.} \label{ssim}
\end{figure}

In Figure \ref{ssim}, we show the PSNR and SSIM scores on the EchoNet-Dynamic test set as a function of the rank $k$, which controls the compression rate. Our \textit{LRM-Functa} approaches (blue and orange curves) is the only method that maintains stable reconstruction quality at very low ranks ($k=2$ and $k=4$). At higher ranks, \textit{LRM-Functa}$_{o}$  aligns with or even outperforms all comparing methods. 

\subsubsection{Frame Detection Results:}
 In Table~\ref{tab:edes}, we report frame detection accuracy on EchoNet-Dynamic for ranks 
$k=2$ and $k=512$, corresponding to compression rates of roughly 3000× and 25×, respectively. \textit{LRM-Functa}$_{b}$
 achieves results close to the state-of-the-art fully supervised approach \cite{li2023semi}, which reports an ED MAE of 2.1 and an ES MAE of 1.7 on the same test set \cite{li2023semi,yang2025latent}. For OOD evaluation on the POCUS test set for $k=512$, all INR-based methods maintain strong frame detection performance, while the convolutional \textit{Autoencoder} fails to generalize.
Overall, the MAE scores on the POCUS dataset are lower than on EchoNet-Dynamic, likely due to the lower frame rate.
\begin{table}[h]
\centering
\caption{Mean $\pm$ standard deviation for the ED and ES MAE on the EchoNet-Dynamic and the OOD POCUS test sets. All methods are unsupervised.}
\label{tab:edes}
\begin{tabular}{lcccccc}
\toprule
 & \multicolumn{2}{c}{EchoNet, $k=2$} 
 & \multicolumn{2}{c}{EchoNet, $k=512$} 
 & \multicolumn{2}{c}{OOD POCUS, $k=512$} \\
\cmidrule(lr){2-3} \cmidrule(lr){4-5} \cmidrule(lr){6-7}
Method 
 & ED MAE 
 & ES MAE 
 & ED MAE  
 & ES MAE 
 & ED MAE 
 & ES MAE  \\
\midrule

\textit{Coin++}
& $8.29 \pm 12.3$ 
& $5.82 \pm 10.8$
& $4.46 \pm 6.9$  
& $3.26 \pm 5.5$ 

&$1.83 \pm 1.1$ & $2.32 \pm 1.4$ \\

\textit{MedFuncta}
& $3.72 \pm 3.8$ 
& $3.48 \pm 3.4$ 
& $2.90 \pm 3.3$ 
& $2.17 \pm 2.4$
&$ 1.61 \pm 1.3$ & $1.27 \pm 1.5$\\

\textit{VidFuncta}
& $3.06 \pm 2.6$ 
& $2.90 \pm 2.8$ 
& $3.08 \pm 3.3$ 
& $2.37 \pm 2.4$
&$\textbf{1.33} \pm \textbf{1.2}$ & $1.89 \pm 3.8$ \\

\textit{Autoencoder}
& $ 2.81\pm3.5$
& $ 2.26\pm2.7$

& $3.82 \pm 4.7$ & $2.44 \pm 2.6$ 
&$3.06 \pm 3.5$ & $4.24 \pm 2.6$ \\

\textit{LRM-Functa}$_{b}$ 
& $\textbf{2.26} \pm \textbf{2.5}$ 
& $\textbf{1.98} \pm \textbf{2.1}$ 
& $\textbf{2.48} \pm \textbf{2.9}$ 
& $2.21 \pm 2.2$
&$1.40 \pm 1.5$ & $\textbf{ 1.13} \pm \textbf{1.4}$ \\

\textit{LRM-Functa}$_{o}$ 
& $2.42 \pm 2.7$ 
& $2.15 \pm 2.2$ 
& $3.04 \pm 3.5$ 
& $\textbf{2.16} \pm \textbf{2.2}$
& $\textbf{1.33} \pm \textbf{1.3}$ & $1.39 \pm 2.5$ \\

\bottomrule

\end{tabular}
\end{table}

\subsubsection{Downstream Task on the Reconstructions:}
 Table \ref{tab:ef_bline} summarizes the peformance of downstream models trained and tested on reconstructed videos $\hat{\mathcal{V}}$. The bottom row shows scores for the original videos $\mathcal{V}$, serving as an upper bound. For ejection fraction, even at a compression level of $k=2$, \textit{LRM-Functa}$_{o}$ largely preserves downstream performance.
This highlights the strong compression capability of our approach while retaining clinically relevant information.
 For the LUS dataset, finer spatial details are needed to reliably detect B-lines, therefore we focus on $k=512$. 
Reconstructions of all \textit{Functa}-based approaches achieve competitive AUROC performance, preserving diagnostically relevant features, whereas the \textit{Autoencoder} fails to retain details required for this task.

\begin{table}[t]
\centering
\caption{Mean $\pm$ std of downstream performance on $\hat{\mathcal{V}}$, evaluated on the reconstructed EchoNet-Dynamic and LUS test sets using 5-fold cross-validation.}
\label{tab:ef_bline}
\setlength{\tabcolsep}{1pt} 
\begin{tabular}{lccc|ccc}

\toprule
 & \multicolumn{3}{c}{Ejection fraction, $k=2$} 
 & \multicolumn{3}{c}{B-line classifier [$\times 10^{-2}$], $k=512$} \\
\cmidrule(lr){2-4} \cmidrule(lr){5-7}
Method 
 & MAE & RMSE & R$^2$ score
 & Accuracy & F1 score & AUROC \\
\midrule

\textit{Coin++} 

&$9.15 \pm 0.1$& $12.17 \pm 0.2$  & $0.03 \pm 0.03$

& 

77.5 $\pm$ 5.0 & 67.8 $\pm$ 8.7 & 80.9 $\pm$ 9.0\\

\textit{MedFuncta} 
& $9.42 \pm 0.1$& $12.94 \pm 0.1$ & $\text{-}0.02 \pm 0.02$ 
& 
81.0 $\pm$ 6.3 & 73.1 $\pm$ 7.7 & 90.1 $\pm$ 1.9\\
\textit{VidFuncta}
& $9.14 \pm 0.1$ & $13.09 \pm 0.2$  & $\text{-}0.02 \pm 0.03$ &
\textbf{83.5} $\pm$ \textbf{5.2} & \textbf{76.6} $\pm$ \textbf{7.3} & \textbf{91.9} $\pm$ \textbf{3.6}\\

\textit{Autoencoder} &

$7.16 \pm 0.3$& $9.79 \pm 0.5$  & $0.47 \pm 0.06$  &
 68.0 $\pm$ 2.1 & 50.2 $\pm$ 6.5 & 66.6 $\pm$ 5.7\\

\textit{LRM-Functa}$_{b}$ 

&$5.70 \pm 0.2$ & $7.63 \pm 0.3$ & $\textbf{0.67} \pm \textbf{0.03}$ &
77.5 $\pm$ 6.4 & 70.9 $\pm$ 3.9 & 89.5 $\pm$ 2.2\\

\textit{LRM-Functa}$_{o}$ 
 & $ \textbf{5.29} \pm \textbf{0.1}$ & $\textbf{7.14} \pm \textbf{0.1}$ & $\textbf{0.67} \pm \textbf{0.01}$ &

82.0 $\pm$ 5.4 & 75.3 $\pm$ 5.6 & 91.7 $\pm$ 3.7\\

\midrule
\textit{Original}
& 4.93 $\pm$ 0.3 
& 6.71 $\pm$ 0.4
& 0.70 $\pm$ 0.03
&  

87.5 $\pm$ 5.6 & 82.8 $\pm$ 7.2 & 91.5 $\pm$ 5.9 \\

\bottomrule
\end{tabular}
\end{table}

\section{Conclusion}

We explore the latent space of time-resolved INR architectures for ultrasound video analysis and introduce \textit{LRM-Functa}, which applies a low-rank constraint to the modulated latent space. This improves reconstruction quality even at extreme compression rates of $k=2$ values per frame and allows direct visualization of the latent space for model interpretability.
Reconstructions and downstream tasks demonstrate that diagnostically relevant information for estimating ejection fraction and classifying B-lines is preserved.
The resulting latent space is clearly structured and interpretable: the cardiac cycle forms a continuous trajectory that can be directly visualized, and ED and ES frames can be identified without additional model training. Our approach outperforms state-of-the-art methods and generalizes to OOD cardiac views. Traversing trajectories in the latent space enables sampling of novel frames, revealing smooth transitions from ED to ES.
In future work, we will leverage this latent representation for further downstream tasks, such as anomaly detection and congenital heart disease classification. We also aim to further improve reconstruction fidelity to preserve fine structural details necessary for assessing left ventricular hypertrophy. Overall, our results highlight the potential of \textit{LRM-Functa} latent spaces to provide compact and explainable representations for medical video analysis.

\begin{credits}
\subsubsection{\ackname}  This study was funded
by the Swiss National Science Foundation (grant number P500PT$\_$222349).

\subsubsection{\discintname}
The authors have no competing interests to declare that are
relevant to the content of this article.
\end{credits}

%
%

%
 \bibliographystyle{splncs04}
 \bibliography{mybibiliography}

@inproceedings{wolleb2025vidfuncta,
  title={VidFuncta: Towards Generalizable Neural Representations for Ultrasound Videos},
  author={Wolleb, Julia and Bieder, Florentin and Friedrich, Paul and Tagare, Hemant D and Papademetris, Xenophon},
  booktitle={International Workshop on Advances in Simplifying Medical Ultrasound},
  pages={109--119},
  year={2025},
  organization={Springer}
}

@inproceedings{
friedrichmedfuncta,
title={MedFuncta: A Unified Framework for Learning Efficient Medical Neural Fields},
author={Paul Friedrich and Florentin Bieder and Julian McGinnis and Julia Wolleb and Daniel Rueckert and Philippe C. Cattin},
booktitle={Medical Imaging with Deep Learning},
year={2026}}

@inproceedings{yang2025latent,
  title={Latent Motion Profiling for Annotation-Free Cardiac Phase Detection in Adult and Fetal Echocardiography Videos},
  author={Yang, Yingyu and Yang, Qianye and Cui, Kangning and Peng, Can and D’Alberti, Elena and Hernandez-Cruz, Netzahualcoyotl and Patey, Olga and Papageorghiou, Aris T and Noble, J Alison},
  booktitle={International Conference on Medical Image Computing and Computer-Assisted Intervention},
  pages={316--325},
  year={2025},
  organization={Springer}
}

@article{dupont2022coin++,
  title={Coin++: Neural compression across modalities},
  author={Dupont, Emilien and Loya, Hrushikesh and Alizadeh, Milad and Goli{\'n}ski, Adam and Teh, Yee Whye and Doucet, Arnaud},
  journal={arXiv preprint arXiv:2201.12904},
  year={2022}
}

@article{dupont2022data,
  title={From data to functa: Your data point is a function and you can treat it like one},
  author={Dupont, Emilien and Kim, Hyunjik and Eslami, SM and Rezende, Danilo and Rosenbaum, Dan},
  journal={arXiv preprint arXiv:2201.12204},
  year={2022}
}

@article{mcginnis2025optimizing,
  title={Optimizing Rank for High-Fidelity Implicit Neural Representations},
  author={McGinnis, Julian and H{\"o}lzl, Florian A and Shit, Suprosanna and Bieder, Florentin and Friedrich, Paul and M{\"u}hlau, Mark and Menze, Bj{\"o}rn and Rueckert, Daniel and Wiestler, Benedikt},
  journal={arXiv preprint arXiv:2512.14366},
  year={2025}
}

@article{sitzmann2020implicit,
  title={Implicit neural representations with periodic activation functions},
  author={Sitzmann, Vincent and Martel, Julien and Bergman, Alexander and Lindell, David and Wetzstein, Gordon},
  journal={Advances in neural information processing systems},
  volume={33},
  pages={7462--7473},
  year={2020}
}

@article{jin2025artificial,
  title={Artificial Intelligence in Ultrasound Imaging: A Review of Progress from Machine Learning to Large Language Model},
  author={Jin, Tong and Yu, Xiaohu and Ai, Zheng and Guo, Hongcheng},
  journal={Advanced Ultrasound in Diagnosis and Therapy},
  volume={9},
  number={4},
  pages={483--496},
  year={2025},
  publisher={PringMa, LLC.}
}

@inproceedings{jiao2020self,
  title={Self-supervised representation learning for ultrasound video},
  author={Jiao, Jianbo and Droste, Richard and Drukker, Lior and Papageorghiou, Aris T and Noble, J Alison},
  booktitle={2020 IEEE 17th international symposium on biomedical imaging (ISBI)},
  pages={1847--1850},
  year={2020},
  organization={IEEE}
}

@misc{takizawa2025videoclipmodelmultiview,
      title={Video CLIP Model for Multi-View Echocardiography Interpretation}, 
      author={Ryo Takizawa and Satoshi Kodera and Tempei Kabayama and Ryo Matsuoka and Yuta Ando and Yuto Nakamura and Haruki Settai and Norihiko Takeda},
      year={2025},
      eprint={2504.18800},
      archivePrefix={arXiv},
      primaryClass={cs.CV},
      url={https://arxiv.org/abs/2504.18800}, 
}

@article{al2024spatiotemporal,
  title={Spatiotemporal Deep Learning-Based Cine Loop Quality Filter for Handheld Point-of-Care Echocardiography},
  author={Al Mukaddim, Rashid and MacKay, Emily and Gessert, Nils and Erkamp, Ramon and Sethuraman, Shriram and Sutton, Jonathan and Bharat, Shyam and Jutras, Melanie and Baloescu, Cristiana and Moore, Christopher L and others},
  journal={IEEE Transactions on Ultrasonics, Ferroelectrics, and Frequency Control},
  volume={71},
  number={11},
  pages={1577--1587},
  year={2024},
  publisher={IEEE}
}

@article{grutman2025implicit,
  title={Implicit neural representation for scalable 3D reconstruction from sparse ultrasound images},
  author={Grutman, Tal and Bismuth, Mike and Glickstein, Bar and Ilovitsh, Tali},
  journal={npj Acoustics},
  volume={1},
  number={1},
  pages={14},
  year={2025},
  publisher={Nature Publishing Group UK London}
}

@article{zhu2025implicit,
  title={Implicit neural representation for medical image reconstruction},
  author={Zhu, Yanjie and Liu, Yuanyuan and Zhang, Yihang and Liang, Dong},
  journal={Physics in Medicine \& Biology},
  volume={70},
  number={12},
  pages={12TR01},
  year={2025},
  publisher={IOP Publishing}
}

@article{hu2022lora,
  title={Lora: Low-rank adaptation of large language models.},
  author={Hu, Edward J and Shen, Yelong and Wallis, Phillip and Allen-Zhu, Zeyuan and Li, Yuanzhi and Wang, Shean and Wang, Lu and Chen, Weizhu and others},
  journal={ICLR},
  volume={1},
  number={2},
  pages={3},
  year={2022}
}

@inproceedings{truong2025low,
  title={Low-Rank Adaptation of Neural Fields},
  author={Truong, Anh and Mahmoud, Ahmed H and Konakovi{\'c} Lukovi{\'c}, Mina and Solomon, Justin},
  booktitle={Proceedings of the SIGGRAPH Asia 2025 Conference Papers},
  pages={1--12},
  year={2025}
}

@inproceedings{perez2018film,
  title={Film: Visual reasoning with a general conditioning layer},
  author={Perez, Ethan and Strub, Florian and De Vries, Harm and Dumoulin, Vincent and Courville, Aaron},
  booktitle={Proceedings of the AAAI conference on artificial intelligence},
  volume={32},
  number={1},
  year={2018}
}

@inproceedings{ryser2022anomaly,
  title={Anomaly detection in echocardiograms with dynamic variational trajectory models},
  author={Ryser, Alain and Manduchi, Laura and Laumer, Fabian and Michel, Holger and Wellmann, Sven and Vogt, Julia E},
  booktitle={Machine Learning for Healthcare Conference},
  pages={425--458},
  year={2022},
  organization={PMLR}
}

@inproceedings{ouyang2019echonet,
  title={Echonet-dynamic: a large new cardiac motion video data resource for medical machine learning},
  author={Ouyang, David and He, Bryan and Ghorbani, Amirata and Lungren, Matt P and Ashley, Euan A and Liang, David H and Zou, James Y},
  booktitle={NeurIPS ML4H Workshop: Vancouver, BC, Canada},
  volume={5},
  year={2019}
}

@article{li2023semi,
  title={Semi-supervised learning improves the performance of cardiac event detection in echocardiography},
  author={Li, Yongshuai and Li, He and Wu, Fanggang and Luo, Jianwen},
  journal={Ultrasonics},
  volume={134},
  pages={107058},
  year={2023},
  publisher={Elsevier}
}

@article{farhad2023cardiac,
  title={Cardiac phase detection in echocardiography using convolutional neural networks},
  author={Farhad, Moomal and Masud, Mohammad Mehedy and Beg, Azam and Ahmad, Amir and Ahmed, Luai A and Memon, Sehar},
  journal={Scientific reports},
  volume={13},
  number={1},
  pages={8908},
  year={2023},
  publisher={Nature Publishing Group UK London}
}

@article{schafer2011savitzky,
  title={What is a savitzky-golay filter?[lecture notes]},
  author={Schafer, Ronald W},
  journal={IEEE Signal processing magazine},
  volume={28},
  number={4},
  pages={111--117},
  year={2011},
  publisher={IEEE}
}

@inproceedings{hara2018can,
  title={Can spatiotemporal 3d cnns retrace the history of 2d cnns and imagenet?},
  author={Hara, Kensho and Kataoka, Hirokatsu and Satoh, Yutaka},
  booktitle={Proceedings of the IEEE conference on Computer Vision and Pattern Recognition},
  pages={6546--6555},
  year={2018}
}

@inproceedings{maiya2024latent,
  title={Latent-inr: A flexible framework for implicit representations of videos with discriminative semantics},
  author={Maiya, Shishira R and Gupta, Anubhav and Gwilliam, Matthew and Ehrlich, Max and Shrivastava, Abhinav},
  booktitle={European Conference on Computer Vision},
  pages={285--302},
  year={2024},
  organization={Springer}
}

@article{mada2015define,
  title={How to define end-diastole and end-systole? Impact of timing on strain measurements},
  author={Mada, Razvan O and Lysyansky, Peter and Daraban, Ana M and Duchenne, J{\"u}rgen and Voigt, Jens-Uwe},
  journal={JACC: Cardiovascular Imaging},
  volume={8},
  number={2},
  pages={148--157},
  year={2015},
  publisher={American College of Cardiology Foundation Washington, DC}
}

@inproceedings{peng2021neural,
  title={Neural body: Implicit neural representations with structured latent codes for novel view synthesis of dynamic humans},
  author={Peng, Sida and Zhang, Yuanqing and Xu, Yinghao and Wang, Qianqian and Shuai, Qing and Bao, Hujun and Zhou, Xiaowei},
  booktitle={Proceedings of the IEEE/CVF conference on computer vision and pattern recognition},
  pages={9054--9063},
  year={2021}
}

@article{bauer2023spatial,
  title={Spatial functa: Scaling functa to imagenet classification and generation},
  author={Bauer, Matthias and Dupont, Emilien and Brock, Andy and Rosenbaum, Dan and Schwarz, Jonathan Richard and Kim, Hyunjik},
  journal={arXiv preprint arXiv:2302.03130},
  year={2023}
}

@article{hernandez2021deep,
  title={Deep learning in spatiotemporal cardiac imaging: A review of methodologies and clinical usability},
  author={Hernandez, Karen Andrea Lara and Rienm{\"u}ller, Theresa and Baumgartner, Daniela and Baumgartner, Christian},
  journal={Computers in Biology and Medicine},
  volume={130},
  pages={104200},
  year={2021},
  publisher={Elsevier}
}

@article{oikonomou2024artificial,
  title={Artificial intelligence-enabled detection and phenotyping of left ventricular hypertrophy on real-world point-of-care cardiac ultrasonography and its implications for patient outcomes},
  author={Oikonomou, Evangelos and Holste, Gregory and Coppi, Andreas and Baloescu, Cristiana and McNamara, Robert and Khera, Rohan},
  journal={Circulation},
  volume={150},
  number={Suppl\_1},
  pages={A4139043--A4139043},
  year={2024},
  publisher={Lippincott Williams \& Wilkins Hagerstown, MD}
}

@inproceedings{byra2024implicit,
  title={Implicit neural representations for speed-of-sound estimation in ultrasound},
  author={Byra, Michal and Jarosik, Piotr and Karwat, Piotr and Klimonda, Ziemowit and Lewandowski, Marcin},
  booktitle={2024 IEEE Ultrasonics, Ferroelectrics, and Frequency Control Joint Symposium (UFFC-JS)},
  pages={1--4},
  year={2024},
  organization={IEEE}
}

@inproceedings{molaei2023implicit,
  title={Implicit neural representation in medical imaging: A comparative survey},
  author={Molaei, Amirali and Aminimehr, Amirhossein and Tavakoli, Armin and Kazerouni, Amirhossein and Azad, Bobby and Azad, Reza and Merhof, Dorit},
  booktitle={Proceedings of the IEEE/CVF International Conference on Computer Vision},
  pages={2381--2391},
  year={2023}
}

@article{wold1987principal,
  title={Principal component analysis},
  author={Wold, Svante and Esbensen, Kim and Geladi, Paul},
  journal={Chemometrics and intelligent laboratory systems},
  volume={2},
  number={1-3},
  pages={37--52},
  year={1987},
  publisher={Elsevier}
}

\end{document}